\documentclass[a4paper,10pt]{article}
\usepackage[utf8]{inputenc}

\newcommand{\vs}{\vspace{0.6em}}

\usepackage{lineno,hyperref}
\modulolinenumbers[5]

\usepackage{booktabs} % For formal tables
\usepackage{enumitem} % to enumerate items
\usepackage{bbm}

\usepackage{tikz}
\usetikzlibrary{shapes,arrows}
\usepackage{graphicx}
\usepackage{amsmath}
\usepackage{amssymb}
\usepackage{amsfonts}
\usepackage{subcaption}

\usepackage{bbm}
\usepackage{dsfont}

\usepackage{algorithm}
\usepackage{algpseudocode}

\usepackage{pgfplots}
\usetikzlibrary{pgfplots.statistics}

\usepackage{booktabs}

\usepackage{mathrsfs}
\def\BIC{\mathrm{BIC}}

\usepackage{xcolor}
\definecolor{lightblue}{RGB}{230,247,254}
\usepackage{enumitem}
\usepackage{soul,color,url}
\usepackage{tabularx}
\usepackage{multirow,bm}
\usepackage{colortbl}

\definecolor{Gray}{gray}{0.85}
\definecolor{LightGray}{gray}{0.95}
\definecolor{LightCyan}{rgb}{0.88,1,1}
\newcolumntype{a}{>{\columncolor{Gray}}c}
\newcolumntype{b}{>{\columncolor{LightGray}}c}
\newcolumntype{g}{>{\columncolor{lightblue}}c}

\DeclareMathOperator*{\argmax}{arg\,max}

\def\LL{\mathrm{LL}}

\def\PL{\mathrm{Pen}}
\def\BIC{\mathrm{BIC}}

\pgfplotsset{
  /pgfplots/xlabel near ticks/.style={
     /pgfplots/every axis x label/.style={
        at={(ticklabel cs:0.5)},anchor=near ticklabel
     }
  },
  /pgfplots/ylabel near ticks/.style={
     /pgfplots/every axis y label/.style={
        at={(ticklabel cs:0.5)},rotate=90,anchor=near ticklabel}
     }
  }

\newcommand{\ktreebuild}[1]{
	\begin{subfigure}[b]{0.4\textwidth}
		  \begin{tikzpicture}[
			scale = 0.5,
			main_node/.style={circle,draw, scale=0.6},
			main_node2/.style={fill=blue!15,draw,ellipse, scale=0.6},
			main_edge/.style={-, line width=1pt}]

	    \node  at (-2.5, 2)  {(#1)};
			
		\node[main_node] (0) at (0, 0)  {A};
		\node[main_node] (1) at (-0.5, 2)  {B};
		\node[main_node] (2) at (0.6, 1.3) {C};

		\draw[main_edge] (0) edge (1);
		\draw[main_edge] (1) edge (2);
		\draw[main_edge] (0) edge (2);

		\node[main_node2] (10) at (3.75, 1.25)  {ABC};
		
		\ifnum #1 > 1

		\node[main_node] (3) at (-1.25, 0.25) {D};
		\node[main_node2] (11) at (5.5, 2.2)  {ABD};
		\draw[main_edge] (10) edge (11);

		\draw[main_edge] (1) edge (3);
		\draw[main_edge] (0) edge (3);
		
		\fi
		
		\ifnum #1 > 2

		\node[main_node] (4) at (1.3, 0.2) {E};

		\node[main_node2] (12) at (5, 0.25)  {ACE};
		\draw[main_edge] (10) edge (12);

		\draw[main_edge] (2) edge (4);
		\draw[main_edge] (0) edge (4);

		\fi
		
		\ifnum #1 > 3
		
		\node[main_node] (5) at (2.1, 1.2) {F};

		\node[main_node2] (13) at (7, 0.3)  {CEF};
		\draw[main_edge] (12) edge (13);

		\draw[main_edge] (2) edge (5);
		\draw[main_edge] (4) edge (5);
		
		\fi

		\end{tikzpicture}
	\end{subfigure}
}

\newcommand{\kgbuild}[1]{
	\begin{subfigure}[b]{0.4\textwidth}
		  \begin{tikzpicture}[
 scale=0.5,
    main_node/.style={fill=green!15,circle,draw, scale=0.6},
    main_node2/.style={circle,draw, scale=0.6},
    main_edge/.style={<-, line width=1pt},
    main_edge2/.style={-, line width=1pt}]
    
    \tikzset{fontscale/.style = {font=\relsize{#1}}}
    
        \node  at (-2.5, 2)  {(#1)};

		\node[main_node] (0) at (0, 0)  {A};
		\node[main_node] (1) at (-0.5, 2)  {B};
		\node[main_node] (2) at (0.6, 1.3) {C};

		\draw[main_edge] (0) edge (1);
		\draw[main_edge] (1) edge (2);
		\draw[main_edge] (0) edge (2);
		
		\node[main_node2] (10) at (5, 0)  {A};
		\node[main_node2] (11) at (4.5, 2)  {B};
		\node[main_node2] (12) at (5.6, 1.3) {C};
		
		\draw[main_edge2] (10) edge (11);
		\draw[main_edge2] (11) edge (12);
		\draw[main_edge2] (10) edge (12);
		
		\ifnum #1 > 1

		\node[main_node] (3) at (-1.25, 0.25) {D};
		\node[main_node2] (13) at (3.75, 0.25) {D};

		\draw[main_edge] (3) edge (1);
		\draw[main_edge] (3) edge (0);

		\draw[main_edge2] (13) edge (11);
		\draw[main_edge2] (13) edge (10);

		\fi

		\ifnum #1 > 2

		\node[main_node] (4) at (1.3, 0.2) {E};
		\node[main_node2] (14) at (6.3, 0.2) {E};

		\draw[main_edge] (4) edge (2);

		\draw[main_edge2] (14) edge (12);
		\draw[main_edge2] (14) edge (10);

		\fi

		\ifnum #1 > 3

		\node[main_node] (5) at (2.1, 1.2) {F};
		\node[main_node2] (15) at (7.1, 1.2) {F};

		\draw[main_edge] (5) edge (2);
		\draw[main_edge] (5) edge (4);

		\draw[main_edge2] (12) edge (15);
		\draw[main_edge2] (14) edge (15);
		
		\fi
		   
		  \end{tikzpicture}
	\end{subfigure}
}

\pgfplotsset{
red/.style={draw=red},
black/.style={draw=black},
}

\title{Efficient Learning of Bounded-Treewidth Bayesian Networks from Complete and Incomplete Data Sets}

\author{Mauro Scanagatta \\ IDSIA, Switzerland \\ mauro@idsia.ch
\and Giorgio Corani \\ IDSIA, Switzerland \\ giorgio@idsia.ch
\and Marco Zaffalon \\ IDSIA, Switzerland \\ zaffalon@idsia.ch
\and Jaemin Yoo \\ Seoul National University \\ jaeminyoo@snu.ac.kr
\and U Kang \\ Seoul National University \\ ukang@snu.ac.kr}

\begin{document}

\maketitle

\begin{abstract}	
Learning a Bayesian networks with bounded treewidth is important for reducing the complexity of the inferences.
We present a novel anytime algorithm (k-MAX) method for this task, which scales up to thousands of variables.
Through extensive experiments we show that it consistently yields higher-scoring structures than its competitors on complete data sets. 
% Structure learning algorithms commonly assume the data set to be complete; yet most common data sets are incomplete.
We then consider the problem of structure learning from incomplete data sets. 
This can be addressed by structural EM, which however is computationally very demanding.
We thus adopt the novel k-MAX algorithm in the maximization step of structural EM, obtaining an efficient computation of the expected sufficient statistics.
We test the resulting structural EM method on the task of imputing missing data, comparing it against the state-of-the-art approach based on random forests. 
Our approach achieves the same imputation accuracy of the competitors, but in about one tenth of the time. 
Furthermore we show that it has worst-case complexity linear in the input size, and that it is easily parallelizable.  
% --- at the best of our knowledge, it is the first approach in the literature able to do so. 

\end{abstract}

\section{Introduction}

The size of an explicit representation of the joint distribution of $n$ categorical random variables is exponential in $n$.
Bayesian networks \cite{darwiche2009modeling} compactly represent joint distributions by exploiting  independence relations 
and encoding them into a directed acyclic graph (DAG), also referred to as  \textit{structure}. 
Yet, algorithms able to perform structure learning 
from thousands of variables
have been devised only very recently for Bayesian networks \cite{scanagatta2015a,scanagatta2016} and for chordal log-linear graphical models (that can be exactly mapped on Bayesian networks) \cite{Petitjean2015-SDM,Petitjean2016-KDD}. 

Given a Bayesian network, the task of computing the marginal distribution of a set of variables, possibly given evidence on another set of variables, is called \textit{inference}.
The complexity of exact inference grows exponentially in the \textit{treewidth} \cite[Chap. 7]{darwiche2009modeling} of the DAG, under the exponential time hypothesis \cite{Kwisthout2010}. 
In order to allow tractable inference we thus need to learn Bayesian networks with a bounded-treewidth structure; this problem is NP-hard \cite{korhonen2013}. 

Most research on learning bounded-treewidth Bayesian networks adopts a score-based approach. 
The score measures the fit of the DAG to the data; 
the goal is hence to find the highest-scoring DAG  that respects the treewidth bound. 
Exact methods \cite{korhonen2013,Parviainen2014,BergJM14} exist, but their applicability is restricted to small domains. 
Approximate approaches that scale up to some hundreds of variables \cite{NieCJ15,nieCJ16} have been more recently proposed.
A recent breakthrough has been achieved by the k-greedy algorithm \cite{scanagatta2016}.
It consistently yields higher-scoring DAGs than its competitors and it scales to several \textit{thousands} of variables.

In this paper we present a new algorithm called k-MAX, which improves over k-greedy.
Both k-MAX and k-greedy are anytime algorithms: they can be stopped at any moment, yielding the current best solution. 
k-MAX adopts a set of more sophisticated heuristics compared to k-greedy;
as a result it consistently  yields higher-scoring DAGs than both k-greedy and other competitors,
as demonstrated by our extensive experiments on complete data sets.

Structure learning algorithms commonly assume data sets to be complete; yet real data sets are often incomplete.
Structure learning on incomplete data sets can be accomplished via 
the \emph{structural expectation-maximization} (SEM) algorithm \cite{Friedman1997}, which alternates between an estimation of the sufficient statistics given the current model (expectation step), and the search of a new model given the expected sufficient statistics (maximization step).
Yet, SEM is computationally demanding:
in particular the expectation step  requires computing several inferences, which might become prohibitive if the model has unbounded treewidth and/or there are many missing data whose
actual value has to be inferred.
We adopt k-MAX as the structure learning algorithm within SEM; 
in this way we obtain a fast implementation of SEM,
since the bounded-treewidth structures learned in the different iterations 
perform efficient inferences. To the best of our knowledge, this is the first implementation of SEM that is able to scale to thousands of variables.
  
To test our method, we use the Bayesian networks learned by SEM in order to perform data imputation.
We consider as a competitor a recent method for data imputation based on random forests \cite{Stekhoven2012} and we compare the two approaches on
data sets with different degrees of missingness.
The two approaches achieve the same imputation accuracy, but our approach 
is faster by almost one order of magnitude. 
Furthermore we show that the complexity of our method scales linearly in the input size (Subsec. \ref{sec:scalability}), and that it is easily parallelizable (Subsec. \ref{subsec:parallelization}).
To the best of our knowledge, it is the first approach in the literature able to do so. 

%===Bring somewhere else%
% The task is  typically performed by implementing a Markov network on the given graph, and performing \textit{approximated} inference through \textit{Loopy Belief Propagation} \cite{Yedidia2003},
% whose complexity is linear time in the number of edges.
% On one hand LBP is not guaranteed to converge, and if it does its solution is approximate; however in practice these approximations are often good.
% In both cases it yields state-of-the-art results on graphs involving several thousands of variables.
% Even using LBP, inference might be too slow if the graph contains many edges
% and the aim is a real-time application such as those previously mentioned.
% We speed up LBP by performing graph subsampling.

In Section \ref{sec:tw} we present the technical background of the paper. In Section \ref{sec:learning} we detail our approach for bounded-treewidth structure learning, k-MAX. In Section \ref{sec:exps} and \ref{sec:chord} we evaluate its performance against existing state-of-the-art approaches. In Section \ref{sec:sem} we present how k-MAX can be used in the SEM algorithm, obtaining the SEM-k-MAX algorithm. It is evaluated in Section \ref{sec:imput} on the task of data imputation against the state-of-the-art approach. Section \ref{sec:concl} concludes our paper. 

The software of this paper is available from \url{http://ipg.idsia.ch/software/blip}, together 
with supplementary material containing the detailed results of our experiments.

\section{Treewidth and $k$-trees}
\label{sec:tw}

Intuitively, the treewidth $k$ quantifies the extent to which a graph resembles a tree.
Following the terminology of \cite{elidan2008} we now provide a formal definition.
Let us recall that a \textit{clique} of an undirected graph is a subset of its nodes such that every two distinct nodes are linked by an edge. 
Moreover, a clique is \textit{maximal} if it is not a subset of a larger clique.

\textit{Treewidth of an undirected graph.}
We denote an undirected graph by $H=(V,E)$ where $V$ is the vertex set and $E$ is the edge set.
% A \textit{tree decomposition} of $H$
% is a pair ($\mathcal{C},\mathcal{T}$) where $\mathcal{C}=\{C_1,C_2,...,C_m\}$
% is a collection of subsets of $V$ and $T$ is a tree on $\mathcal{C}$, so that:
% \begin{itemize}
% 	\item  $ \cup_{i=1}^{m} \,\, C_i=V$;
% 	\item for every edge which
% 	connects the vertices $v_1$ and $v_2$,
% 	there is a subset $C_i$ which contains both $v_1$ and $v_2$;	
% 	\item for all $i,j,k$ in $\{1,2,..m \}$ if $C_j$ is on the  path between
% 	$C_i$ and $C_k$ in $\mathcal{T}$ then $C_i \cap C_k  \subseteq C_j$.
% \end{itemize}
% The width of  a tree decomposition is $\max(|C_i|)-1$ where
% $|C_i|$ is the number of vertices in $C_i$.
% The treewidth of $H$ is the minimum width
% among all possible tree decompositions of $G$.	
	An undirected graph is \textit{triangulated} when every cycle of length greater than or equal to 4 has a \textit{chord}, that is, an edge connecting two non-consecutive nodes in the cycle \cite[Def. 9.16]{darwiche2009modeling}.  Triangulated graphs are also called  \textit{chordal} graphs. 
The \textit{triangulation} of a graph 
is the operation of adding chords until the graph is triangulated. 
The treewidth of a triangulated graph is the size of its largest clique minus one.
The treewidth of  $H$ is the minimum treewidth among all the possible triangulations of $H$.

\textit{Treewidth of a Bayesian network.}
The moral graph of the DAG  
associated to a Bayesian network 
is an undirected graph that includes an edge ($i$ -- $j$) for every edge ($i\rightarrow j$) in  the DAG and an edge ($p$  -- $q$) for every pair of edges ($p\rightarrow i$), ($q\rightarrow i$) in the DAG.
The treewidth of the DAG is the treewidth of its moral graph.

\subsection{$k$-trees}
\begin{figure*}[!ht]
	\centering
	
	\includegraphics{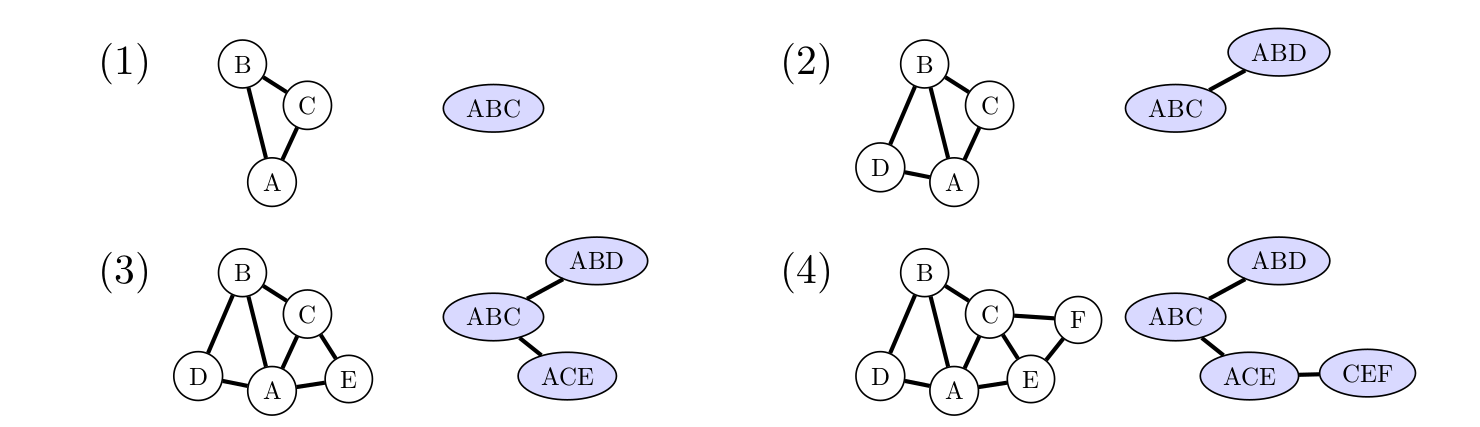}

	\caption{Iterative construction of
		a $k$-tree (k = 2). We start with the clique over the nodes \{A,B,C\}; then we add the nodes D, E and F one at a time.
		When we add a node, we link it to a 2-clique of the existing graph.
		In purple  we show the  decomposition of the graph into its maximal cliques.
		All maximal cliques have size
		three; thus the treewidth of the graph is two.
		\label{fig:ktree-build}
	}
\end{figure*}

A $k$-tree is an undirected \textit{edge-maximal} graph of treewidth $k$, that is,  the addition of any edge to the $k$-tree increases its treewidth.
It is defined inductively as follows \cite{patil1986}. 
\textit{Base case}: a clique with $(k+1)$ nodes is a $k$-tree.
\textit{Inductive step}: given a $k$-tree $H_n$ on $n$ nodes, a $k$-tree $H_{n+1}$ on $(n+1)$ nodes is obtained by connecting 
the $(n+1)$-th node to a $k$-clique of $H_n$ (a $k$-clique is a clique over $k$ nodes).
See Figure \ref{fig:ktree-build} for an example.
As a final remark, a sub-graph of a $k$-tree is called  \textit{partial $k$-tree}; its treewidth is at most $k$.

\section{Structure learning of Bayesian networks}
\label{sec:learning}

We consider the problem of learning the structure of a Bayesian network from a complete data set.
The set of $n$ categorical random variables  is $\mathcal{X}=\{X_1, ..., X_n\}$. The goal is to find the highest-scoring bounded-treewidth DAG $\mathcal{G} = (V, E)$, where $V$ is the collection of nodes and $E$ is the collection of arcs. $E$ can be represented by the set of parents ${\Pi_1, ..., \Pi_n}$ of all variables.

Structure learning is usually accomplished in two steps. First,
\textit{parent set identification} is the identification of
a list (\textit{cache}) $L_i$ of candidate parent sets
independently for each variable  $X_i$. 
Second, \textit{structure optimization} is the assignment
of a parent set to each node in order to maximize the score of the resulting
DAG.

The problem of bounded-treewidth structure learning
can be casted as follows:
\begin{align*}
& \mathcal{G}^* =
\argmax score(\mathcal{G})\ \mathrm{s. t.}\\	
& \forall_i: \Pi_i\in L_i\\
& \mathcal{G}\mathrm{~is~DAG} \\
& \mathrm{treewidth}(\mathcal{G}) \leq k . \\
\end{align*}

Different scores can be used to assess the fit of a DAG.
We adopt the {\it Bayesian information criterion} ($\mathrm{BIC}$),
which is asymptotically proportional to the posterior probability of the DAG.
The $\mathrm{BIC}$ score is defined as follows:
\begin{align}
& \BIC(\mathcal{G})  = \sum_{i=1}^{n} \BIC(X_i,\Pi_i)=\sum_{i=1}^n
\left(\LL(X_i|\Pi_i) + \PL(X_i,\Pi_i)\right) \; ,
\end{align}
where $\LL(X_i|\Pi_i)$ denotes the log-likelihood of $X_i$ and its parent set:
\begin{align}
& \LL(X_i|\Pi_i) =\displaystyle\sum\nolimits_{\pi \in \Pi_i, ~ x \in
	X_i} N_{x,\pi}\log\hat{\theta}_{x|\pi} \; ,
\end{align}
while $\PL(X_i,\Pi_i)$ is the \textit{complexity penalization}:
\begin{align}
& \PL(X_i,\Pi_i) = -\frac{\log N}{2}(|X_i|-1)(|\Pi_i|) \; .
\end{align}
% \begin{align}
% & \mathrm{BIC}(\mathcal{G})=\displaystyle \sum\nolimits_{i=1}^{n} \mathrm{BIC} (X_i, \Pi_i) =\\
% & \displaystyle \sum\nolimits_{i=1}^{n} \displaystyle \sum\nolimits_{\pi \in |\Pi_i|}  \displaystyle \sum\nolimits_{x \in |X_i|} N_{x,\pi}\log\hat{\theta}_{x|\pi}- \frac{\log N}{2}(|X_i|-1)(|\Pi_i|)\, ,
% \end{align}
We denote by 
$\hat{\theta}_{x|\pi}$ the maximum likelihood estimate of the conditional probability $P(X_i=x|\Pi_i=\pi)$; 
by $N_{x,\pi}$ the number of times that $(X=x\land\Pi_i=\pi)$ appears in the data set; $|\cdot|$ indicates the size of the Cartesian product space of the variables given as argument.
Thus $|X_i|$ is the number of states of $X_i$ and $|\Pi_i|$ is the product
of the number of states of the parents of $X_i$.

The BIC score is \textit{decomposable}, namely it is constituted by the sum of the scores of the individual variables given their parents. 
The k-MAX algorithm, which we present later, can be applied to any decomposable scoring functions;
see \cite{Liu2012} for a discussion of decomposable scoring functions.

\subsection{Parent set identification}
In order to efficiently prepare the cache $L_i$ of candidate parent sets of each variable $X_i$
we adopt the approach of  \cite{scanagatta2015a}.
The main idea of~\cite{scanagatta2015a} is to quickly identify the most promising parent sets
through an approximate scoring function that does not require scanning the data set.  
%Later on,
%only the scores of the most promising parent sets are computed. 
The approximate scoring function is
called $\mathrm{BIC}^*$.  The $\mathrm{BIC}^*$ of a parent set $\Pi = \Pi_{1} \cup \Pi_{2}$
constituted by the union of two non-empty and disjoint parent sets $\Pi_{1}$ and $\Pi_{2}$ is:
\begin{equation}
\mathrm{BIC}^*(X, \Pi_{1}, \Pi_{2}) = \mathrm{BIC}(X,\Pi_{1}) +
\mathrm{BIC}(X,\Pi_{2}) + \mathrm{inter}(X, \Pi_{1}, \Pi_{2})\, ,
\end{equation}
\noindent
that is, the sum of the $\mathrm{BIC}$ scores of the two parent sets and of an interaction term,
which ensures that the penalty term of $\mathrm{BIC}^*(X, \Pi_{1}, \Pi_{2})$ matches the penalty
term of $\mathrm{BIC}(X, \Pi_{1} \cup \Pi_{2})$.  
In particular,
%$\mathrm{inter}(X, \Pi_{1}, \Pi_{2}) = \frac{\log N}{2}(|X|-1)(|\Pi_{1}| + |\Pi_{2}| - |\Pi_{1}|
%|\Pi_{2}| - 1) - \mathrm{BIC}(X,\emptyset)$.  The $\mathrm{BIC}^*(X,\Pi)$ score is equal to the
%$\BIC(X,\Pi)$ score if the interaction information $ii(X; \Pi_{1}; \Pi_{2})$ is
%zero~\cite{scanagatta2015a}.  Yet, this condition is generally false; for this reason,
%$\mathrm{BIC}^*(X,\Pi)$ is an \textit{approximate} score, but it is efficiently computable.
%
If $\mathrm{BIC}(X,\Pi_{1})$ and $\mathrm{BIC}(X,\Pi_{2})$ are known, then $\mathrm{BIC}^*$ is
computed in constant time (with respect to data accesses).
The \textit{independence selection} algorithm~\cite{scanagatta2015a} exploits $\mathrm{BIC}^*$ to
quickly approximately score a large number of parent sets
without limiting the in-degree, which is the maximum number of parents allowed for every node.
Eventually, it computes the actual  score of the most promising parent sets.
Additionally we adopt pruning rules \cite{deCampos2009} in order to avoid computing the score of sub-optimal parent sets.

\subsection{Learning Bayesian networks with bounded treewidth}
Exact methods for bounded-treewidth structure learning of Bayesian networks
\cite{korhonen2013,Parviainen2014,BergJM14}  scale to at most a few dozens of variables.
Approximate approaches are therefore needed to scale to larger domains.

The  S2 algorithm \cite{NieCJ15}
uniformly samples the space of $k$-trees; then it assesses the sampled $k$-trees through a heuristic scoring function (\textit{informative score}).
The DAG is then obtained by constraining its moral graph to be a sub-graph of the
$k$-tree with highest informative score.
The S2+ algorithm \cite{nieCJ16} further refines this idea, obtaining via A* the $k$-tree guaranteed to maximize the informative score. In general S2+ recovers higher-scoring DAGs than S2 but its scalability is limited: for instance it cannot be used with
thousands of variables.

\subsubsection{k-greedy}
To the  best of our knowledge, the state-of-the-art algorithm for bounded-treewidth learning is
so far constituted by \textit{k-greedy} \cite{scanagatta2016}, which consistently yields higher-scoring DAGs than its competitors.
The k-greedy algorithm samples the space of the orderings of variables, and given an ordering it
builds the bounded-treewidth DAG inductively as follows.

\textit{Initialization.} Given an order over the variables and the value of $k$,
k-greedy initializes the structure with a DAG over the first $(k+1)$ variables in the order.
The DAG is learned using either the exact method of \cite{cussens11} 
or the approximate method of \cite{scanagatta2017}, depending on the value of $(k+1)$ .
The treewidth of the learned DAG is at most $k$ (its moral graph is a sub-graph of the clique over the same $k+1$ variables, which has treewidth $k$).

%\textit{Feasible parent set given the $k$-tree.}
%Let us denote by $k$ the bound on the treewidth set by the user.
%We define a parent set $\Pi_i \in L_i$ as \emph{feasible given the  current $k$-tree} iff $ \exists k_j \in \mathcal{K}_C \mathrm{s. t.} \Pi_i \subseteq k_j $, where $\mathcal{K}_C$ is the set of existing $k$-clique in $\mathcal{K}$.

\textit{Addition of the following nodes.}
Given a DAG $\mathcal{G}_q$ over $q$ variables, k-greedy chooses the highest-scoring \emph{feasible} parent set for the variable $(q+1)$ in the order, $X_{\prec q+1}$.
A parent set is feasible if it is a $k$-clique (or a subset of a $k$-clique) of the moral graph of $\mathcal{G}_q$.
Once the parent set of $X_{\prec q+1}$ is chosen, we obtain the DAG $\mathcal{G}_{q+1}$ over  $q+1$ variables.
At each iteration the moral graph of the DAG is a partial $k$-tree: thus the treewidth of the  DAG is bounded by $k$.

Then k-greedy samples a new ordering and repeats the above procedure. 
When a maximum number of iterations or a maximum execution time is met it returns the highest-scoring DAG found. 

\subsection{k-MAX}

\begin{figure*}[!ht]
	\centering
	
	\includegraphics{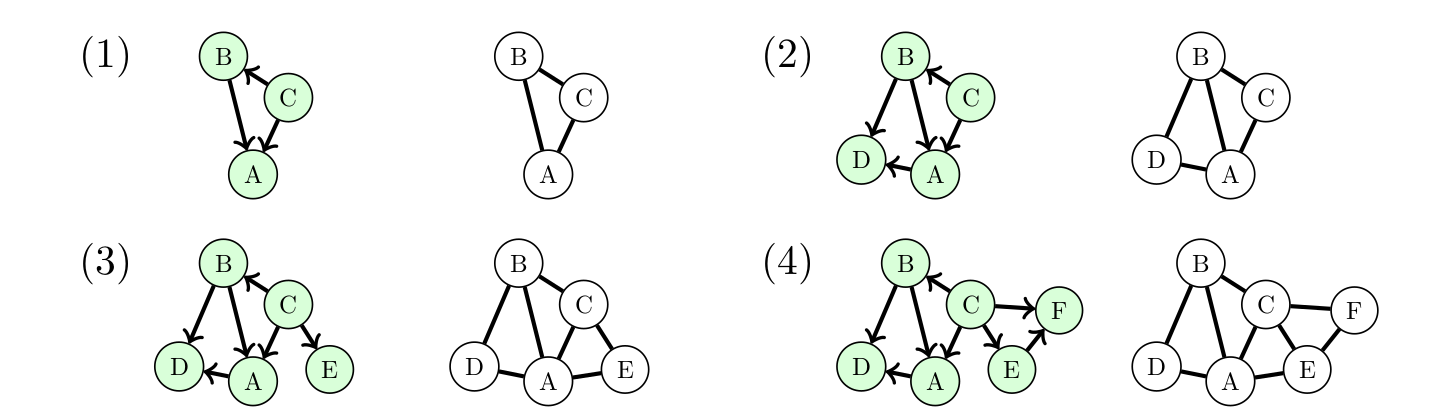}
	
	\caption{
		Example of a treewidth-bounded ($k$=2) DAG  built  by k-MAX. For each iteration we also show the underlying k-tree graph (white nodes), which allows to check that the treewidth is indeed bounded by $k$=2. At iteration (2), $m(D)>m(E)$
		and $m(E)>m(F)$; hence node $D$ is inserted. Its highest-scoring feasible parent set is constituted by \{A,B\}.
		At iteration (3) we still have  $m(E)>m(F)$ and thus we insert $E$ . Its highest-scoring parent set is constituted by only \{C\}, so in the k-tree we choose a k-clique  that is
		a random superset of \{C\}.  At iteration (4) we insert the last remaining variable $F$. 
		\label{fig:kg-build}
	}
\end{figure*}	

k-MAX shares a fundamental idea with k-greedy: it
incrementally grows a DAG, guaranteeing that at each step its
moral graph is a sub-graph of a $k$-tree.
Differently from k-greedy, k-MAX
does not adopt a predefined ordering over the variables.
Instead it ranks  the variables that can be inserted in the graph
through the heuristic score $m(X_i)$:
\begin{equation*}
	m(X_i) = \frac{sc^C(X_i) - sc^W(X_i)}{sc^B(X_i) - sc^W(X_i)} \; ,
\end{equation*}
where:
\begin{align*}
	sc^C(X_i) = \max_{\Pi \in L^*_i} score(\Pi) \; , \\
	sc^B(X_i) = \max_{\Pi \in L_i} score(\Pi) \; , \\
	sc^W(X_i) = \min_{\Pi \in L_i} score(\Pi) \; .
\end{align*}

$L^*_i$ is the subset of parent sets that are \emph{feasible}. Recall that the feasible parent sets
are constituted by the $k$-cliques of the moral graph of the current DAG, and by subsets of such $k$-cliques.

The $m(X_i)$ heuristic compares for each variable the highest-scoring \emph{feasible} parent set 
with the lowest-scoring and the highest-scoring parent set available in the cache $L_i$ (notice that most parent sets in $L_i$ are not feasible
when the $k$-tree contains a small number of variables).
The rationale is  to defer the addition of variables whose
$m(\cdot)$ score is low, as it will increase in the subsequent iterations due to availability
of a larger number of feasible parent sets.	
The scores
$sc^B(X_i)$ and $sc^W(X_i)$ can be found in linear time w.r.t. $|L_i|$ and cached. As for
$sc^C(X_i)$, it needs to be updated each time a new variable is added to the DAG.

We present an outline of k-MAX in Alg. \ref{alg:kmax}. 
The procedure is repeated until a specific termination condition is met (for example maximum execution time). 
The highest-scoring DAG found is then returned. 

\begin{algorithm}
\caption{k-MAX}
\label{alg:kmax}
\begin{algorithmic}[1]
\Procedure{k-MAX}{$\mathcal{X}$ // Set of variables}
\While {not termination condition}
\State $ q \gets k+1 $
 \State $\mathcal{G}_{q} \gets Initialization(\mathcal{X})$ // Learn the initial graph with $k+1$ variables
 \While {$q < |\mathcal{X}|$}
 \State $\mathcal{G}_{q+1} \gets Addition(\mathcal{G}_{q})$ // Add the variable with the highest $m(X_i)$ score
 \State $ q \gets q + 1 $
 \EndWhile
 \EndWhile
 \State // Return the highest-scoring $\mathcal{G}$ found
\EndProcedure
\end{algorithmic}
\end{algorithm} 

%k-MAX first samples $k+1$ variables from which it builds a starting graph, and proceeds to add the variable with the highest $m(X_i)$ until all are inserted in the graph. At that point a bounded-treewidth DAG over all the variables is complete. 
% As in k-greedy it then repeats the above procedure until a pre-defined condition is met (maximum number of iterations or a maximum execution time), and returns the highest-scoring DAG found. 

\paragraph{Initialization}
We start by building  an initial $k$-tree $\mathcal{K}_{k+1}$ over $k+1$ variables as follows.
We initialize the list of chosen variables $\mathcal{I}$ as an empty set, and we create a set $\mathcal{C}$, which stores the \emph{candidate} parents: namely every variable appearing at least in one parent set of $X_i \in \mathcal{I}$.
We first choose randomly the first variable $X$, we add it to the set $\mathcal{I}$ and we add its candidate parents to $\mathcal{C}$.
Then, until $\mathcal{I}$  contains $k+1$ variables we: (1) add to it a new variable $Z$ chosen randomly from $\mathcal{C}$; (2) add all the candidate parents of $Z$ to $\mathcal{C}$.

Finally we learn the initial DAG  $\mathcal{G}_{k+1}$ over the variables contained in $\mathcal{I}$, either with exact or approximate methods, as in the initialization of k-greedy. 
The moral graph of  $\mathcal{G}_{k+1}$ is a sub-graph of $\mathcal{K}_{k+1}$ and thus $\mathcal{G}_{k+1}$ has  treewidth at most $ k$.
For each new $k$-clique added to $\mathcal{K}_{k+1}$, we update $m(X_i)$ for each variable $X$ not yet processed.

\paragraph{Addition of the following nodes}
Let us denote by  $\mathcal{G}_{q}$ and $\mathcal{K}_{q}$  the current DAG and the $k$-tree over $q$ nodes.

The $(q+1)$-th variable to be added is:
$$X = \argmax_{Z \notin G} m(Z) \; .$$
We connect $X$ to the parent set:
$$\Pi^C_{X} = \argmax_{\Pi \in L^*_i} score(\Pi) \; .$$
This yields the updated DAG $\mathcal{G}_{q+1}$.
We then update the $k$-tree, connecting $X$ to the $k$-clique that is superset of $\Pi^C_{X}$.
In the event of several $k$-cliques sharing this property, a random one between them is chosen. 
This yields the $k$-tree $\mathcal{K}_{q+1}$; it contains an additional ($k+1$)-clique compared to $\mathcal{K}_{q}$.
By construction, $\mathcal{K}_{q+1}$ is also a $k$-tree.
For each new $k$-clique added to $\mathcal{K}_{q+1}$, we update $m(X_i)$ for each variable $X_i$ not yet processed.

\paragraph{Space of learnable DAGs}
A reverse topological order is an order ${X_1 , ...X_n }$ over the vertices of a DAG in which each vertex $X_i$
appears before its parents $\Pi_i$. The search space of k-MAX contains only DAGs whose
reverse topological order, when used as variable elimination order, has treewidth $k$.
The proof is identical to that provided in \cite{scanagatta2016}
for k-greedy.
The extensive experiments  by \cite{scanagatta2016} show that such limitation does not hurt the empirical performance of k-greedy and thus we do not further discuss this point.

\subsection{Advantages over k-greedy}
\paragraph{Initialization}

Recall that k-MAX builds iteratively the initial set of ($k+1$) variables by ensuring that any added variable is a candidate parent of at least another variable already in the clique.

We prune the available parent sets for each variable by Lemma 1 of \cite{decampos2011a}.
It states that given two parent sets $\Pi^1_i$ and $\Pi^2_i$ for the same variable $X_i$, such that $\Pi^2_i \subseteq \Pi^1_i$ and
$\mathrm{score}(\Pi^2_i) < \mathrm{score}(\Pi^1_i)$,
then $\Pi^1_i$ can be discarded from $L_i$ as it yields sub-optimal structures.
This happens when $\Pi^1_i$ has low mutual information with $X_i$;
its contribution to the score is negative since the small increase in log-likelihood (due to low mutual information) is outweighed by
the complexity penalization of the scoring function.
Thus a variable is a \emph{candidate parent} only if it appears in at least one non-pruned parent set of $X_i$
(and which thus has significant mutual information with $X_i$).

Instead the initialization of k-greedy randomly samples $k+1$ variables, ignoring their mutual information. 
% On both cases the highest-scoring DAG over the $k+1$ variables is found, but 
By virtue of the selection process the initial DAG found by k-MAX is in general higher-scoring than the one found by k-greedy.

%Let us assume that the score function employs a penalization factor based on the size of the parent set.
%This is necessary, since otherwise the graph with the maximal score would simply be the complete one, leading to ineviTable~ overfitting.
% This results in a higher-scoring initial DAG compared to k-greedy.

\paragraph{Addition of the following nodes}

The addition step of k-greedy follows the order, which is randomly sampled.
Again, this overlooks whether there exists a feasible parent set given the current $k$-tree with a good score  for the variable being added.
k-MAX instead optimizes the variable to be added at each iteration, by ranking them according to the $m(\cdot)$ score.

\section{Experiments: k-MAX against k-greedy}
\label{sec:exps}

\begin{table}[!ht]
	\begin{center}
			\caption{The 18 real data sets used in the experiments ($n$ is the number of variables, $d$ is the number of data points).}
		\begin{tabular}{cgg| cgg| cgg|}
			\toprule
			Name & $n$ & $d$ & Name & $n$ & $d$ & Name & $n$ & $d$ \\
			\midrule
Kdd & 64 & 11490 & 		Retail & 135 & 4408 & 		EachMovie & 500 & 591 \\
Plants & 69 & 3482 & 		Pumsb-star & 163 & 2452 & 	WebKB & 839 & 838 \\
Audio & 100 & 3000 & 		DNA & 180 & 1186 &		Reuters-52 & 889 & 1540\\
Jester & 100 & 4116 & 		Kosarek & 190 & 6675 &  	C20NG & 910 & 3764 \\
Netflix & 100 & 2634 & 		MSWeb & 294 & 5000 & 		BBC & 1058 & 330 \\
Accidents & 111 & 2551 & 	Book & 500 & 1739 &		Ad & 1556 & 491 \\
			\bottomrule
		\end{tabular}
		\label{tab:data-datasets}
	\end{center}
	\vspace{-2em}
\end{table}

\begin{table}[!ht]
	\begin{center}
			\caption{The 20 known networks from which we sampled synthetic data sets ($n$ is the number of variables).}
		\begin{tabular}{cg cg cg}
			\toprule
			Name & $n$ & Name & $n$ & Name & $n$ \\
			\midrule
andes & 223 & r2 & 2000 & r9 & 4000 \\
diabetes & 413 & r3 & 2000 & r10 & 10000 \\
pigs & 441 & r4 & 2000 & r11 & 10000 \\
link & 724 & r5 & 4000 & r12 & 10000 \\
munin & 1041 & r6 & 4000 & r13 & 10000 \\
r0 & 2000 & r7 & 4000 & r14 & 10000 \\
r1 & 2000 & r8 & 4000 &  & \\
			\bottomrule
		\end{tabular}
		\label{tab:data-networks}
	\end{center}
% 	\vspace{-2em}
\end{table}

We consider the 18 data sets listed in Table~\ref{tab:data-datasets}.
They have been previously
used in \cite{rooshenas2014learning} and in other works referenced therein.
The data sets are available for instance from \url{https://github.com/arranger1044/awesome-spn#dataset}.
Each data set is split in three subsets; we thus perform  18$\cdot$3=54 structure learning experiments.
The number  of instances in each dataset ranges from 226 to 180000. 

Additionally,
we compared k-MAX and k-greedy 
on 20 synthetic data sets sampled from known networks (Table \ref{tab:data-networks}).
Five of these are taken from the literature\footnote{\url{http://www.bnlearn.com/bnrepository/}} ($andes$, $diabetes$, $pigs$, $link$, $munin$) while the
other fifteen (r0-r14) have been generated by us, using the BNgenerator package\footnote{\url{http://sites.poli.usp.br/pmr/ltd/Software/BNGenerator/}}.
They  contain up to 10,000 variables. From each known network we sample a training data set of 5000 instances.

We hence consider a total of 74 data sets (54 real ones and 20 synthetic ones), on which we compare
k-greedy and k-MAX. We provide both algorithms with the same cache of parent sets for each variable, pre-computed using
independence selection \cite{scanagatta2015a}; we then let
each algorithm run for one hour on the same machine.

In each experiment we measure the difference between the BIC scores ($\Delta$BIC) of the DAG returned by k-MAX and k-greedy.
There is an exact mapping between the values of  $\Delta$BIC and the Bayes factor (BF) \cite[Sec. 4.3]{raftery1995bayesian}.
The BIC score of graph $\mathcal{G}$ is an approximation of the logarithm of the marginal likelihood of $\mathcal{G}$,
namely $P(D|\mathcal{G}) = \int P(D|\mathcal{G}, \theta) p(\theta) d\theta \simeq \mathrm{BIC}(\mathcal{G})$.
Given two graphs $\mathcal{G}_1$ and $\mathcal{G}_2$, the Bayes factor (BF) is the ratio of their marginal likelihoods. The log of the Bayes factor can be approximated by the difference  of the BIC scores:
$$ \log(\mathrm{BF}) = \log \left( \frac{P(D|\mathcal{G}_1)}{P(D|\mathcal{G}_2)}  \right)\simeq \mathrm{BIC}(\mathcal{G}_1) - \mathrm{BIC}(\mathcal{G}_2)
= \Delta \mathrm{BIC}.$$ 
A positive $\Delta \mathrm{BIC}$ provides evidence in favor of $\mathcal{G}_1$ and a negative 
$\Delta \mathrm{BIC}$ provides evidence in favor of $\mathcal{G}_2$.
The posterior probability of $\mathcal{G}_1$ is given by 
$$
P(\mathcal{G}_1|D) = \frac{P(D|\mathcal{G}_1)}{P(D|\mathcal{G}_1) + P(D|\mathcal{G}_2)} \simeq
\frac{\exp(\mathrm{BIC}(\mathcal{G}_1))}{\exp(\mathrm{BIC}(\mathcal{G}_1)) + \exp(\mathrm{BIC}(\mathcal{G}_2))}
$$
For instance a $\Delta \mathrm{BIC}$ \textgreater 10 implies a Bayes factor \textgreater 150 and a posterior probability $P(\mathcal{G}_1|D)$ \textgreater 0.99, conveying very strong evidence in favor of $\mathcal{G}_1$. 
Following the same logic, the values of $\Delta$BIC can be interpreted \cite[Sec. 4.3]{raftery1995bayesian}  according  to this scale:
\begin{itemize}
	\item $\Delta$BIC \textgreater 10: extremely positive evidence;
	\item  6 \textless $\Delta$BIC \textless 10: strongly positive evidence;
	\item 2 \textless  $\Delta$BIC \textless 6: positive evidence;
	\item $\Delta$BIC  \textless 2: neutral evidence.
\end{itemize}
We report only the case of positive $\Delta$BIC; negative values of $\Delta$BIC are interpreted in the same way but they have  the meaning of negative evidence.

\begin{table}[!ht]
%	\rowcolors{4}{lightblue}{white}
	\begin{center}
			\caption{
			Comparison between k-greedy and k-MAX under various treewidths. For each treewidth we perform 74 structure learning experiments. The results in favor of k-MAX are statistically significant for each tested treewidth.}
		\begin{tabular}{c c c c}
\toprule & \multicolumn{3}{c}{treewidth} \\
%\midrule
k-MAX vs k-greedy & $2$ & $5$ & $8$   \\
\midrule
$\Delta$BIC & & &  \\
extremely positive & \textbf{71} & \textbf{67} & \textbf{69}  \\
strongly positive & 0 & 0 & 0 \\
positive & 0 & 0 & 1  \\
neutral& 0 & 0 & 0  \\
negative & 0 & 0 & 0  \\
strongly negative & 0 & 0 & 0 \\
very negative & 3 & 7 & 5 \\
			\bottomrule
		\end{tabular}
		\label{tab:kg-kmax}
	\end{center}
% 	\vspace{-2em}
\end{table}
We perform independent experiments with the treewidths $k \in \{2,5,8\}$.
We summarize the results in Table~\ref{tab:kg-kmax}.
In most cases there is an extremely positive evidence for the model learned by
k-MAX over the model learned by k-greedy ($\Delta$BIC \textgreater 10).
We further analyze the results through a sign-test, considering one method as winning over the other when there is  a $\Delta$BIC
of at least 2 in its favor, and treating as ties the cases in which $|\Delta \mathrm{BIC}|$ \textless 2.
The number of wins obtained by k-MAX over k-greedy is significant \textit{for every tested treewidth}.

% 
% \begin{table}[]
% \centering
% \caption{Log-likelihood}
% \label{my-label}
% \begin{tabular}{llll}
% Dataset	    & N   & K-MAX  & k-G \\
% nltcs       & 16  & \textit{-5.32}  & -5.42  \\
% plants      & 69  &  \textit{-14.79} & -15.07 \\
% kdd         & 64  &  \textit{-1.76}  & -1.76  \\
% baudio      & 100 &  \textit{-42.9}  & -42.95 \\
% jester      & 100 &  \textit{-56.44} & -56.52 \\
% bnetflix    & 100 &  \textit{-59.04} & -59.22 \\
% accidents   & 111 & \textit{ -29.07} & -29.52 \\
% tretail     & 135 & -10.75 &  \textit{-10.72} \\
% pumsb\_star & 163 &  \textit{-26.62} & -27.95 \\
% dna         & 180 &  \textit{-79.86} & -80.97 \\
% kosarek     & 190 &  \textit{-11.11} & -11.19 \\
% msweb       & 294 &  \textit{-10.04} & -10.09 \\
% tmovie	    & 500 &  \textit{-61.10} & -62.16 \\
% book 	    & 500 & -37.35 &  \textit{-37.32} \\
% cwebkb 	    & 839 & \textit{-162.20} & -162.70 \\
% cr52 	    & 889 & \textit{-92.48} & -94.63 \\
% c20ng 	    & 910 & \textit{-164.23} & -162.24 \\
% bbc 	    & 1058 &  \textit{-261.09} &  -261.31 \\
% ad	    & 1556 & \textit{-18.59} & -19.59
% \end{tabular}
% \end{table}

\paragraph{Iteration statistics}
We further compare k-MAX and k-greedy by analyzing their iterations.
We consider as an example the data set \emph{tmovie.test} (500 variables) with treewidth $k$=5.
As shown in
Table~ \ref{tab:comparison},
k-MAX performs much less iterations (two orders of magnitude less, in this example) than k-greedy; this is due to the overhead of
updating the $m(\cdot)$ values for all the variables not yet added to the structure.
However this strategy pays off, as  
the median score of the DAG retrieved at each iteration is much higher for k-MAX. 
This is the advantage of using the more sophisticated heuristics of k-MAX.

\begin{table}[!ht]
	\begin{center}
			\caption{Statistics about the execution of k-greedy and k-MAX on the \emph{tmovie.test} dataset (500 variables).}
		\begin{tabular}{gcc}
			\toprule
& k-MAX & k-greedy \\  \midrule
Number of iterations & 1,111 & 96,226 \\
% Mean & -37034.30 & -37406.99  \\
Median BIC score & -36,937 & -37,489 \\
\bottomrule
		\end{tabular}		
		\label{tab:comparison}
	\end{center}
	\vspace{-2em}
\end{table}

\paragraph{Comparison against S2}
In \cite{scanagatta2016} it was shown that k-greedy
consistently outperforms S2.
k-MAX further increases the gap over S2.
In particular k-MAX achieves
$\Delta$BIC \textgreater 10 compared to S2
for every treewidth and data set considered.

\paragraph{Inference times}
We perform some tests about inference times, using
Iterative Join Graph Propagation \cite{mateescu2010join} as inference engine.
We focus first on the networks containing 1,000 or more variables provided in Table \ref{tab:data-networks},
in which case the ground-truth networks are known.
Using such large ground-truth networks results in slow inference even when computing
 \textit{marginals}.
In several cases we had no convergence of the inference even after 30 minutes  of
computation.
In these cases, even if we could manage to learn the actual DAG with a perfect learner, the model would be hardly usable due to the slowness of the inference.
Conversely,
the bounded-treewidth models learned by k-MAX compute marginals consistently
in less than 0.1 seconds, even with treewidth 8.
 By bounding the treewidth we thus guarantee the efficiency of the inferences.
  
Similar considerations hold also for the smaller ground-truth networks, such as andes, diabetes, etc. In these cases marginals can be efficiently computed using the ground-truth networks, but slowness problems appear when we compute the probability
of the \textit{joint} evidence of \textit{five} variables.
This requires (averaging over data set) about 60 seconds when using the ground-truth networks and
 less than 5 seconds when using bounded-treewidth models with treewidth \textit{eight} (the \textit{slowest} bounded-treewidth model).

\section{Comparison with Chordalysis}
\label{sec:chord}

Chordalysis \cite{Petitjean2013-ICDM} performs
structural learning for log-linear models; such undirected graphical models
are  also known as Markov networks.
In particular Chordalysis learns
chordal models, which are at the intersection between Bayesian networks and Markov networks:
given a chordal Markov network it is possible to obtain a 
Bayesian network which encodes exactly the same independences \cite[Sec. 4.5.3]{koller2009}.

Chordalysis starts from the empty graph, which contains no edges.
It then decides which edges to add based on a series of statistical test of independence.
While classical  approaches  to  log-linear  analysis
hardly scale beyond ten variables,
Chordalysis 
scales to high-dimensional data thanks to
sophisticated algorithms for  efficiently computing  the statistics of the test and
avoiding the computation of unnecessary tests.
We learn the undirected chordal graph using the variant
of Chordalysis  referred in \cite{Petitjean2016-KDD} (code available from \url{https://github.com/fpetitjean})
and we subsequently obtain the equivalent Bayesian network using the algorithm of \cite[Sec. 4.5.3]{koller2009}.

While score-based structural learning aims at maximizing the predictivity of the model, structural learning based on hypothesis tests aims at building an explanatory model, by controlling the rate of false positive edges among those which constitute the graph.
Thus such two approaches have different goals.
Yet, it does make sense to compare k-MAX and
Chordalysis. These are among the very few methods able to learn PGMs from thousands of variables; moreover,
even if Chordalysis  does not formally bound the treewidth, it generally
yields quite sparse graph that are likely to have low treewidth (see the results shown later).
This happens because
the statistical test (corrected for multiple comparisons) does not allow adding an arc unless there is strong evidence
against the null hypothesis of independence.

\subsection{Results}

In this comparison we avoid using the BIC score of the resulting networks as a performance indicator, as Chordalysis does not aim at maximizing it. 
Instead, we measure how well the models fit the distribution by computing the log-likelihood of the instances 
of the test set,
$LL= \sum_{i=1}^{k} log (P(D_i|M))$,
where $D_1,\ldots,D_k$ denote the $k$ instances of the test set and $M$ is the model being tested.
We then compute the difference in test-set log-likelihood ($\Delta$LL) between the model learned by k-MAX and  by Chordalysis.
A value greater than 0 indicates that the model learned by k-MAX yields higher likelihood (thus better fit) than the model learned by Chordalysis, and vice versa. 

\begin{figure}[!ht]
	\centering

	 \begin{subfigure}[b]{1\textwidth}
	\centering

		\includegraphics{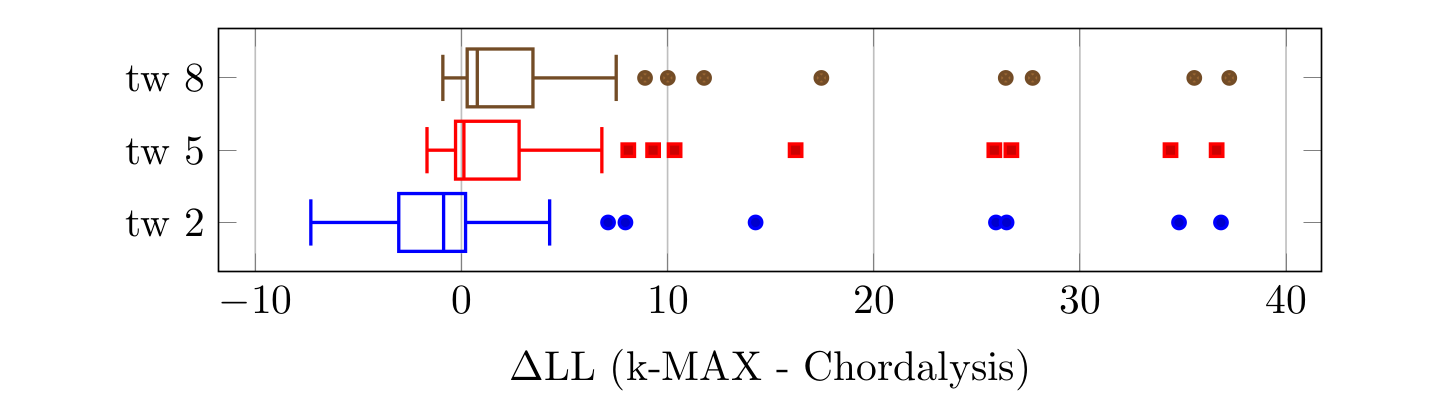}
\subcaption{Real data sets}
\end{subfigure}

\vs

	 \begin{subfigure}[b]{1\textwidth}
	 		\centering
	\includegraphics{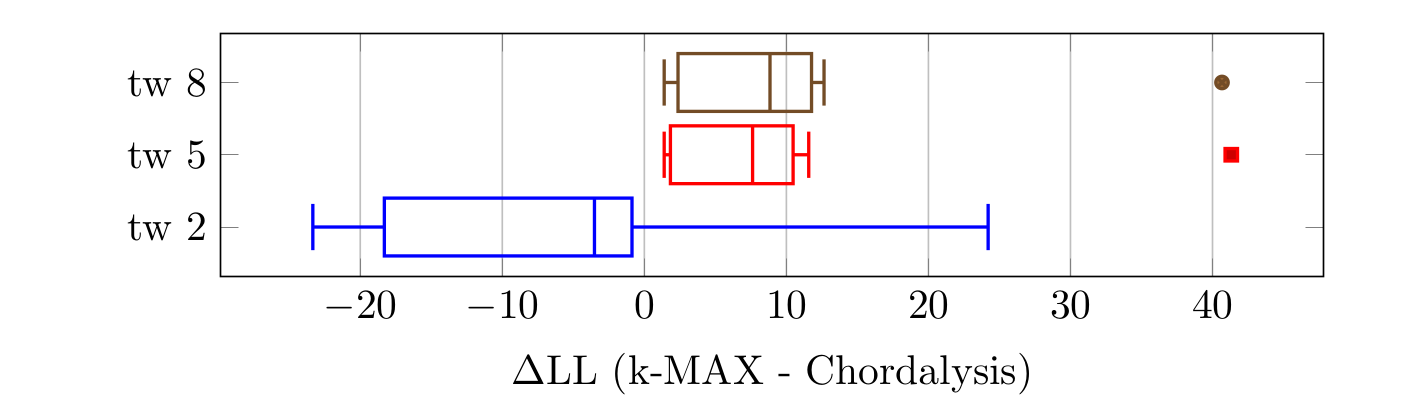}
 	\subcaption{Synthetic data sets}
\end{subfigure}

	\caption{Difference in test-set log-likelihoods between
		the models learned by k-MAX (using various treewidths) and Chordalysis, across the real data sets of Table \ref{tab:data-datasets} and the synthetic data sets of Table \ref{tab:data-networks}.}
	\vspace{-1em}
	\label{fig:dataset-ll}
\end{figure}

We first consider the datasets listed in Table~ \ref{tab:data-datasets}.
Recall that each data set is split into three parts; for each dataset, we use a part for learning the models and the union of the other two parts as a test set.
We thus perform 3 experiments for each data set, for a total of 54 experiments.
In Fig. \ref{fig:dataset-ll} (upper plot) we show the distribution of $\Delta$LL across the data sets for different tested treewidths.
The models learned by Chordalysis provide a fit that is comparable to the models learned by k-MAX using treewidth $2$.
When k-MAX adopts higher treewidths, such as 5 or 8, it fits better (or even much better) the distribution than Chordalysis.

We then consider the true networks of Table \ref{tab:data-networks}. 
In Fig. \ref{fig:dataset-ll} (lower plot) we see that the same pattern appears for the $\Delta$LL.
In this case we sample from the networks $5000$ instances as the training set and $50000$ instances for the test set.

%\begin{figure}[!ht]
%	\centering
%
%		\begin{tikzpicture}
%		\begin{axis}[
%		width=0.90\textwidth,
%		height=0.25\textwidth,
%		grid=major,
%		mark size=1.5pt,
%		ymajorgrids=true,
%		xlabel=$\Delta$LL,
%		%boxplot/draw direction=y,
%		% xtickmin=1, %suppress xticklabel
%	    % xtickmax=0,
%		yticklabels={,,tw 2, tw 5, tw 8},
%		]
%
%		\addplot+[boxplot, thick] table[y index=1] {data/networks-ll.txt};
%		\addplot+[boxplot, thick] table[y index=2] {data/networks-ll.txt};
%		\addplot+[boxplot, thick] table[y index=3] {data/networks-ll.txt};
%		
%		\end{axis}
%		\end{tikzpicture}
%% 	
%	\caption{Distribution of the $\Delta$LL between
%		the models learned by k-MAX (using various treewidths) and Chordalysis, across the data sets of Table \ref{tab:data-networks}.
%		The log-likelihood are computed on the \textit{test sets}.
%	}
%	\vspace{-1em}
%	\label{fig:networks-ll}
%\end{figure}

\begin{figure}[!ht]
	\centering

	\includegraphics{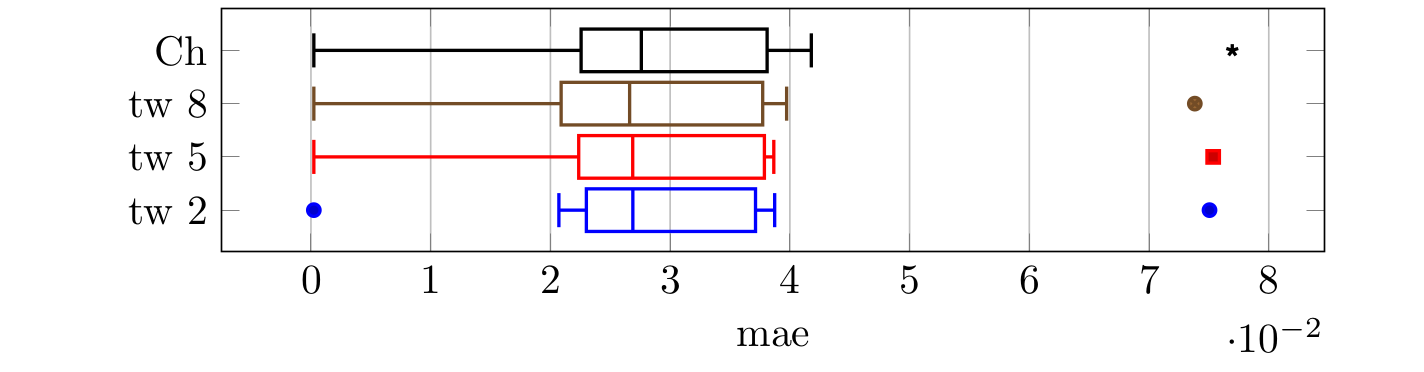}
	\caption{Comparison of the mae of the inference (probability of the evidence of five variables) for the models yielded 
		by Chordalysis and k-MAX under various treewidths. }
	\vspace{-1em}
	\label{fig:networks-mae}
\end{figure}

\begin{figure}[!ht]
	\centering

	\includegraphics{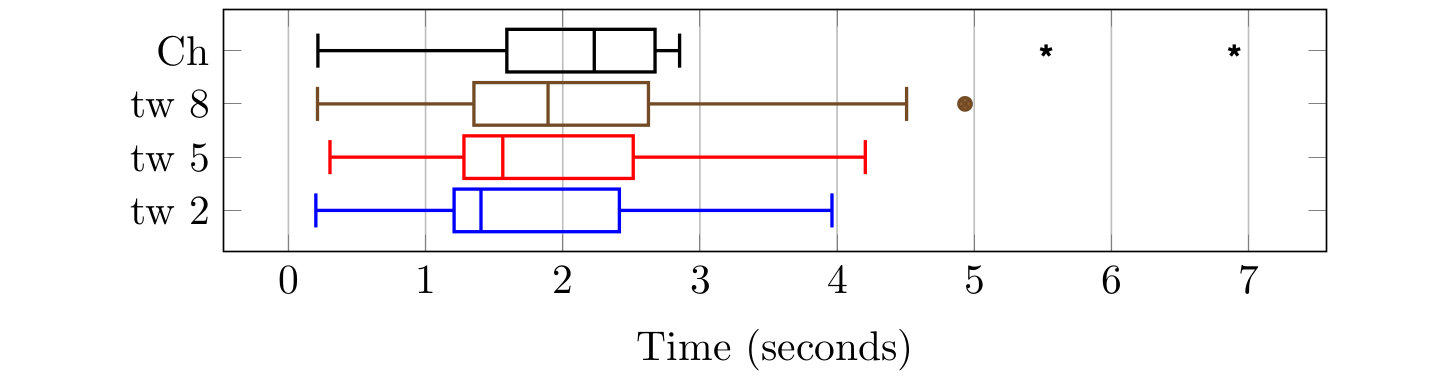}
	\caption{Time required to compute the probability of the joint evidence of five variables using the networks learned by Chordalysis and k-MAX with various treewidths.}
	\vspace{-1em}
	\label{fig:networks-times}
\end{figure}

Additionally we study  the running time and the accuracy of the inference.
As inference we consider the computation of the probability of the evidence $P (e)$ constituted by \emph{five} randomly selected variables, which we set in random states.
For each data set we run 100 queries.
On each data set we measure the mean absolute error (mae) of each model:
\begin{equation*}
\mathrm{mae} = \frac{1}{q} \sum_i | P_i(e)-\hat{P}_i(e) | \; ,
\end{equation*}
where $q$ denotes the total number of queries, $P_i(e)$ and $\hat{P}_i(e)$ the probability of evidence computed  by respectively the ground-truth model and the bounded-treewidth model on the i-th query.
As ground-truth we take the probability of evidence computed on the original network, using the algorithm of Iterative Join Graph Propagation \cite{mateescu2010join} and running it until convergence.

We show in Fig. \ref{fig:networks-mae} the mae of the model learned by Chordalysis and the models learned by k-MAX with the treewidth bound in $\{2, 5,8\}$.
The mae of the learned models are comparable, with a slight improvement with higher value of treewidth (not visible in the graph).
In Fig. \ref{fig:networks-times} we show the time required to compute the probability of evidence.
The inference times of the models learned by k-MAX increase with the treewidth.
Yet even the models learned by k-MAX with treewidth 8 are on average slightly faster than
the model learned by Chordalysis.
Generally the k-MAX models with treewidth 5 or 8 yield both
a better fit
and  a  quicker inference
than Chordalysis.
This shows the soundness of k-MAX.
We recall that however Chordalysis aims at building explanatory models rather than predictive ones. 

\section{Structural EM}
\label{sec:sem}

Most works in structure learning assume  the data set to be \emph{complete}. 
Yet this is rarely the case in real-word domains.
This poses a serious problem, since
learning a structure from \emph{incomplete} data is computationally challenging.

The most common approach for structure learning from incomplete data sets is the  \emph{structural EM} (SEM),
originally presented in \cite{friedman1996learning,friedman1998}; a discussion about this algorithm is given also
by \cite[Chapter ~19]{koller2009}. 
%It augments the \emph{expectation-maximization} algorithm \cite{Dempster1977}, which estimates the parameters from incomplete samples, with an additional step of  structure learning. 

The key idea of structural EM is 
to use the current best estimate of the distribution to complete the data, and then 
analyze such complete data.
In a nutshell, structural EM performs search in the joint space of structure and parameters. 
It alternates between finding better parameters for the current structure, or select a new structure. The former case is as a parametric EM step, while the latter step is a structural EM step. 
For penalized likelihood scoring functions, such as the BIC, this procedure is proven to converge to a local maximum \cite{friedman1996learning}.

More in detail,
structural EM proceeds 
 by alternating two steps. In the \emph{expectation} step, it computes the \emph{expected} sufficient statistics given a candidate structure (the sufficient statistics are
 the counts of the occurrences of each value of a given variable jointly with each possible assignment of its parents). 
Given the expected sufficient statistics,
the \emph{maximization} step learns 
an updated structure and estimates its parameters. The two steps are alternated until the search converges to a structure. We present SEM in Alg.~\ref{alg:sem}, adopting the description of \cite[Chapter ~19]{koller2009}.
 
\begin{algorithm}
\caption{Structural EM algorithm}
\label{alg:sem}
\begin{algorithmic}[1]
\Procedure{Structural-EM}{\par $\mathcal{G}^0$ // Initial Bayesian network structure, \par $\boldsymbol\theta^0 $ // Initial set of parameters for $\mathcal{G}^0$, \par $\mathcal{D}$ // Partially observed data set \par}
\For {each $t = 0, 1, ..., $ until convergence}
\State // Optional parameter learning step
\State  $\boldsymbol{\theta^{{t'}}} \gets $Expectation-Maximization($\mathcal{G}^t, \boldsymbol{\theta^t}, \mathcal{D}$)
 \State // Run EM to generate expected sufficient statistics;
\State // this  yields  the imputed data $\mathcal{D}^*_{\mathcal{G}^t, \boldsymbol\theta^{{t'}}}$.
 \State $\mathcal{G}^{t+1} \gets $ Structure-Learn($\mathcal{D}^*_{\mathcal{G}^t, \boldsymbol\theta^{{t'}}}$)
 \State $\boldsymbol{\theta^{t+1}} \gets $ Estimate-Parameters($\mathcal{D}^*_{\mathcal{G}^t, \boldsymbol\theta^{{t'}}}, \mathcal{G}^{t+1}$) 
\EndFor
% \State return $\mathcal{G}^{t}, \theta^{t}$
\EndProcedure
\end{algorithmic}
\end{algorithm}

The most demanding part of SEM is the expectation step, which requires computing several queries, whose complexity is (in the worst case) exponential in the treewidth of the model.
Thus SEM becomes prohibitive if the model being learned has unbounded treewidth and there are many missing data.
This prevents the application of SEM to large data sets.
%The lack of an effective solution to this problem has prevented the application of the SEM framework on datasets with more than a dozens of variables. 
We adopt k-MAX in order to perform structure learning within SEM:
by learning bounded-treewidth models 
we obtain an efficient computation of the expected sufficient statistics.
We call the resulting approach as \emph{SEM-kMAX}. As far as we know this is the first implementation of the structural expectation-maximization that scales up to thousands of variables. 

The \texttt{Structure-Learn} step of Algorithm~\ref{alg:sem} is 
constituted by two parts.
Given the (expected) sufficient statistics computed in the previous iteration, 
we first compute the cache of best parent sets for each variable using \emph{independence selection}. Then we find the highest-scoring DAG using k-MAX with a treewidth bound of $k=6$.
%Both  algorithms are anytime; as such, they require to set a maximum execution time.  
We set an execution time of $n$ seconds (one second per variable)
for the first step and of $n/10$ seconds
for the second step. 
Such time limits are shorter than those adopted on complete data sets, as we need to perform structure learning at each maximization step of SEM.
In Section~\ref{subsec:tuning} we evaluate the sensitivity of the results on the allowed execution time of both steps. 

The algorithm reaches convergence when the structure remains unchanged between the subsequent SEM iterations and
%We start the structure learning by first
%evaluating the best structure found in the past iteration. Hence, convergence is reached when the structure learning step does not improve 
hence no improvements are found on the structure of the previous iteration.  
As for the choice of the initial network we adopt a random chain that connects all the variables, as in \cite{Friedman1997}.

\subsection{Further implementation details}

%We present now our implementation of the SEM algorithm. 
A peculiar aspect of our implementation is that we adopt the \emph{hard EM} \cite[Chap.~19.2.2.6]{koller2009}
for the computation of the expected sufficient statistics. 
While the standard \emph{soft} EM produces a  probability distribution over the missing data, 
hard EM fills-in 
the missing data with their most probable completion.
The relative merits of hard and soft EM are discussed for instance by
\cite{samdani2012unified} and \cite[Chap.~19.2.2.6]{koller2009}. 
We adopt the hard EM in order to limit the memory usage of the algorithm.
In fact 
we cannot foresee which sufficient statistics will be required 
by the maximization step, when it looks for a new structure.
Keeping the soft-completed data set  in memory is however not feasible:
the memory requirement is, for each instance, 
exponential in the number of missing values.
Some workarounds exist \cite[Chap.~19.4.3.4]{koller2009}, based on severe restrictions on
the type of learnable structures.
By adopting the hard EM we radically solve the memory issues, as the hard-completed data set requires the same memory of the original data set.
This allows us to perform structure learning without restricting the search space of the structures, apart from the bounded-treewidth constraint.

Given an instance of the dataset containing missing values, we can fill the missing values \textit{jointly} or \textit{independently}. 
The joint approach requires running a single MPE (most probable explanation) query  \cite[Chap.2.1.5.2]{koller2009} for each incomplete instance.
The independent approach requires running a marginal query for each missing value, marginalizing out over the missing variables in the same instance.
In the field of multi-label classification (where one has to predict a set of related labels, given the observed features) it has been pointed out 
\cite{cheng2010bayes} that the MPE inference maximizes the probability of correctly predicting 
the \textit{whole} set of labels, while 
the marginal inference maximizes the probability of correctly predicting 
each label independently.
In practical terms, extensive experiments in multi-label classification do not show major  differences between the results yielded by the two approaches \cite{alessandro2013ensemble}.
There are however applications, such as message decoding over a noisy channel, where the joint approach is clearly preferable. 
In the following, we report results obtained using the joint approach.

\section{Application to data imputation}
\label{sec:imput}

We benchmark the performance of SEM-kMAX in the task of \textit{data imputation}, which is the process of replacing missing data with the predictions of their values.
Once we run  SEM until convergence, we have both a trained model and an \textit{imputed} data set.
% The strategy of substituting each missing value with the most common value of the variable (breaking ties at random) is the baseline competitor. 
% We call this approach \emph{mode imputation}.

A pioneering Bayesian network approach for data imputation is that of \cite{di2004bayesian}, which
however requires to order the variables according to their reliability before performing structure learning; this approach is hardly applicable
in data mining applications, where the number of variables can be in order of the thousands.
As a state-of-the-art competitor for data imputation we thus consider an approach based on random forests.

The \textit{missForest} algorithm \cite{Stekhoven2012}  recasts the  problem of missing data imputation as a prediction problem. 
The initial guess is made using mode imputation, namely by substituting missing values with the most common value of the variable. The variables are then sorted according to the amount of missing values and  the data are imputed by regressing (using random forest) each variable in turn against all other variables (starting from the variable with the smallest missingness). The predictions of missing data for the dependent variable are used as imputation.
The empirical study of \cite{Fei2017} compares different imputation algorithms based on random forests, concluding that \textit{missForest} is indeed the most accurate, but also the slowest. 
The problem is that at each iteration it requires fitting $n$ random forests (one for each variable); this becomes slow when dealing with large number of variables.
Thus \cite{Fei2017} propose the mRF$_{\alpha}$ algorithm
as a scalable alternative to \textit{missForest}. It randomly divides the $n$ variables into mutually exclusive groups of approximate size $\alpha p$. Each group in turn acts as the multivariate response to be regressed on the remaining $(1-\alpha)p$ variables at each iteration. Thus at each iteration it trains $1/\alpha$ random forest models.
%the algorithm \emph{missForest} \cite{Stekhoven2012} that was shown to achieve better performance in the imputation of missing data  compared to other approaches based on random forest \cite{Fei2017}. 
%It casts the imputation problem as a prediction problem; the imputations are obtained by regressing each variable in turn against all other variables.
%For each variable $X_s$ (starting from the one with lowest amount of missingness), a random forest is first trained over its observed values and then used for imputing its missing values. The imputation procedure is repeated until a stopping criterion is met. 
%Given $n$ variables, for each iteration $n$ random forest models must be fit, which can be slow in certain domains. 
 We used the implementation of  mRF$_{\alpha}$ available in the \textit{randomForestSRC} R package, 
%  which has been called \emph{mForest} by the authors in \cite{Fei2017}. 
% It is a computationally faster version, able to achieve a 10-fold reduction time with a similar imputation accuracy. 
setting $\alpha=0.25$, which yields good results in the experiments of \cite{Fei2017}. 
%We used the implementation available from the R package \textit{randomForestSRC}.
% As a na\"ive competitor for large datasets we consider the simplest possible application of the BNs structure learning to the SEM algorithm. 
% In the maximization step we constraint the DAG to have in-degree equal to 1. 
% We thus can bound the complexity of the expectation step, as the treewidth of the DAG will be 1. 
% We denote this algorithm as \emph{SEM-tree}.
\subsection{Experimental setup}

We ran imputation experiments on the 18 data sets of Table~\ref{tab:data-datasets}. 
On each dataset we induced missing completely at random (MCAR) missingness: namely, each observation was made missing with  fixed probability, regardless of the values of the other variables. 
For each data set we considered the missingness percentages $\{1, 2, 3, 5, 8, 10, 12, 15 \}$ (meaning, e.g., that in the first setting we made missing each value with probability $1\%$), and we repeated each experiment 5 times. 
In each repetition we measured the 
proportion of missing values that were correctly imputed
(\textit{imputation accuracy}):
\begin{align*}
\mathrm{imputation \, accuracy} = \frac{1}{n}
\sum_{j=1}^{n} \frac{\sum_{j=1}^{m_j} \mathbbm{1}(X^{orig}_{i,j} = X^{imp}_{i,j}) }{m_j} \; ,
\end{align*}
where $X^{orig}_{i,j}$ denotes the value of the variable $X_i$ in instance $j$ in the original dataset,
$X^{imp}_{i,j}$ 
its imputed value, $m_j$ the number of values missing in instance $j$ and $n$ is the total number of instances.

\subsection{Comparison with mRF$_{\alpha}$}

  \begin{figure}[!ht]
\centering 

\includegraphics{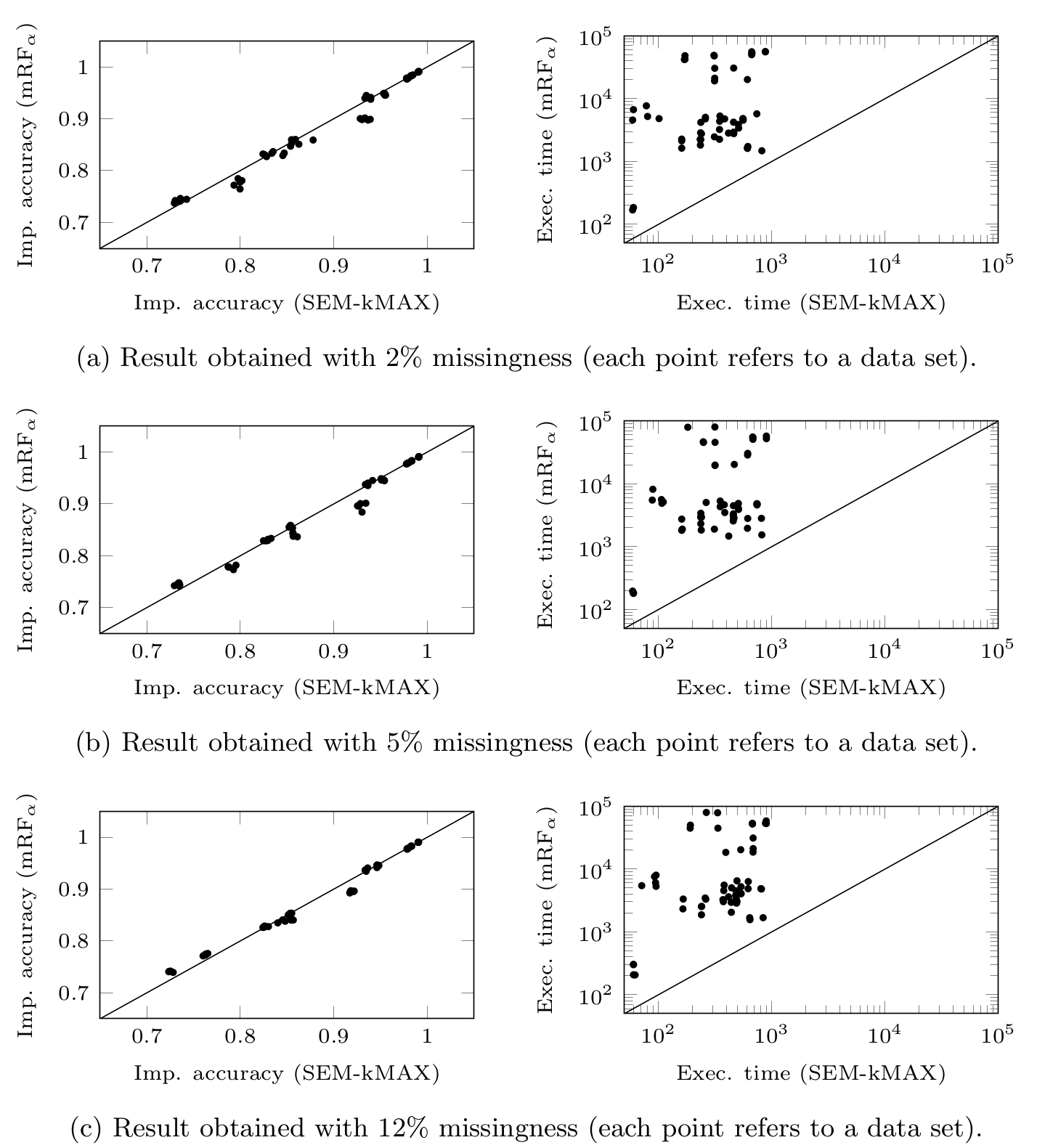}
    	\caption{Comparison on imputation accuracy (on the left) and on execution times (on the right, in seconds) between imputation executed with mRF$_{\alpha}$ and SEM-kMAX.}

    	\label{fig:mainscatter}
\end{figure}

In the comparison with mRF$_{\alpha}$ we used all the available datasets of Table~\ref{tab:data-datasets}.
In two cases (\emph{MSWeb} and \emph{C20NG}) mRF$_{\alpha}$ failed to provide a solution in less than 24 hours, and the datasets were removed from the comparison. 
We conjecture that this is due to the presence of both a high number of variables and a high number of data points. 

Figure~\ref{fig:mainscatter} shows the scatter plots of imputation accuracy and execution times, which compare SEM-kMAX and mRF$_{\alpha}$ for different missingness levels. 
The two approaches offer practically the same imputation accuracy (left plots), but SEM-kMAX is substantially faster than mRF$_{\alpha}$ (right plots).

\begin{figure}[!ht]
\centering
\includegraphics{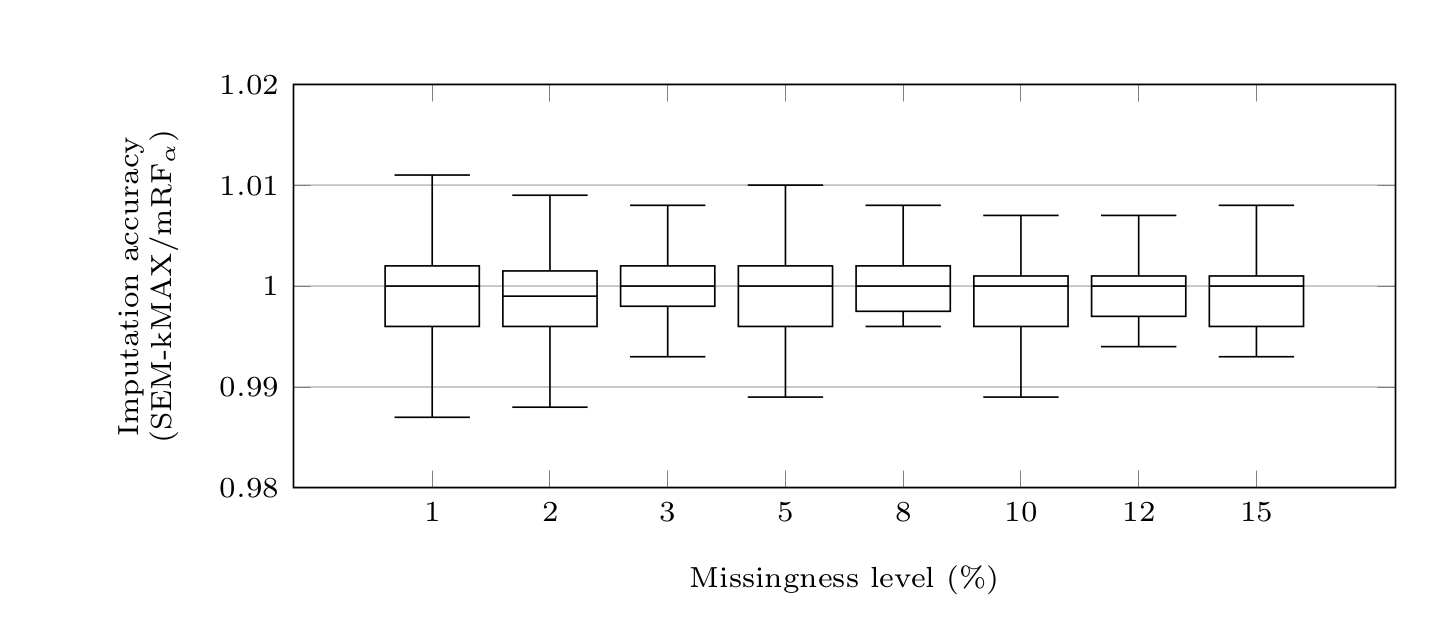}

	 \caption{Aggregate comparison of the imputation accuracy ratio between SEM-kMAX and mRF$_{\alpha}$  imputation.}
	 	 \label{fig:imp1}
 \end{figure}
 
  \begin{figure}[!ht]
\centering
\includegraphics{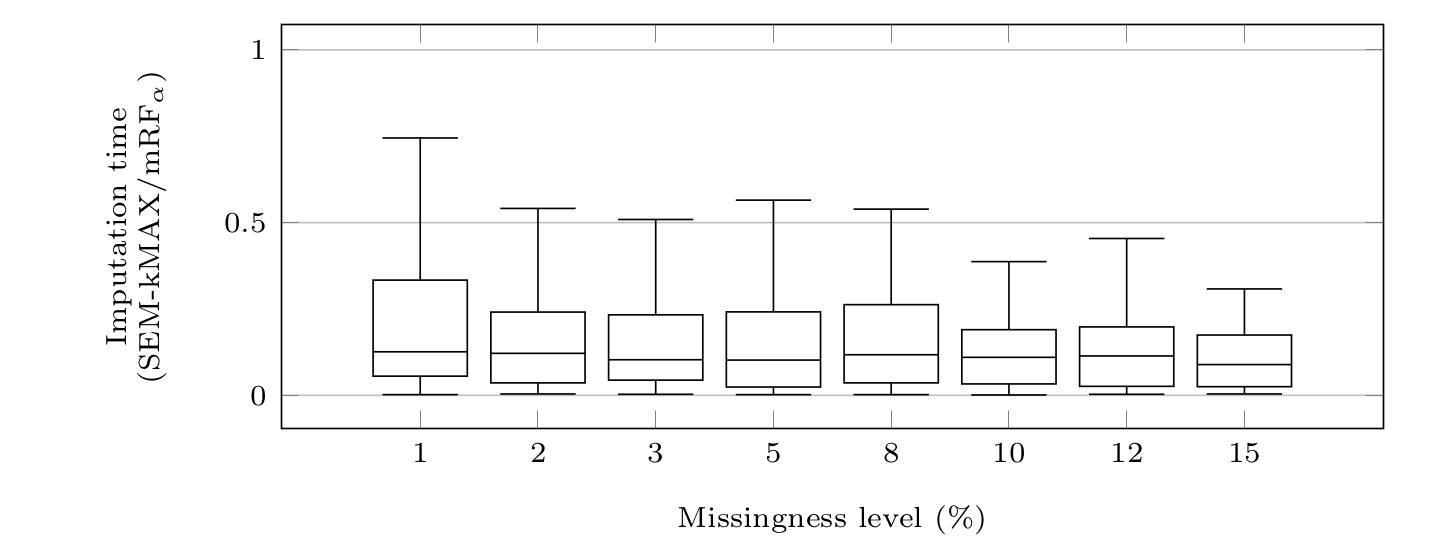}
	 
	 \caption{Aggregate comparison of the execution time ratio between SEM-kMAX and mRF$_{\alpha}$ imputation.}
	 \label{fig:times}
 \end{figure}	

 We further report their ratio of imputation accuracy in Fig.~\ref{fig:imp1} and their ratio of execution times in Fig.~\ref{fig:times} by aggregating  the experiments performed on various data sets but having the same level of missingness. 
 Again we can see that SEM-kMAX achieves an imputation accuracy comparable to mRF$_{\alpha}$, while on average reducing the computational time of one order of magnitude.
 
%  \subsection{Comparison with mode imputation}
% 
%    \begin{figure}[!ht]
% \centering
% 	\begin{subfigure}[b]{0.9\textwidth}
% 	  \res{data/res/aggr/res}{2}{
% 	  	Imputation  accuracy \\ 
% 	  	( SEM-kMAX/mode)
% %	  	$\mathcal{R}_{kM, st}$
%   	}
% 	\end{subfigure}
% 
% 	 
% 	 \caption{Aggregate comparison of the ratio of imputation accuracy between SEM-kMAX and mode imputation.}
% \label{fig:imp2}	 
%  \end{figure}
% 
% %Let $M_i$ be the set of instances in which the value of the variable $X_i$ has been made missing,  $X^{orig}_{i,j}$ the value of the variable $X_i$ at instance $j$ in the original dataset $\mathcal{D}$, $X^{imp}_{i,j}$ the value assigned in the imputed dataset $\mathcal{I}$, and $I(x,y)$ the indicator function which gives 1 if $x=y$ and 0 otherwise. 
% 
% We now compare SEM-kMAX and the baseline offered by mode imputation. For this comparison we used all the datasets, ranging up to 1556 variables. 
% To the best of our knowledge, SEM-kMAX is the first 
% implementation of structural EM that scales to such large data sets. 
% 
% We report in Fig.~\ref{fig:imp2} the ratio of imputation accuracy between SEM-kMAX and mode imputation,
% % it being the only existing approach for handling learning from datasets with thousands of variables. 
% over which SEM-kMAX achieves a substantial advantage, especially when the missingness is limited. 
% However the difference decreases as the missingness increases, as extracting meaningful relationships from the data becomes more and more difficult. 

 \subsection{Computational complexity}
 \label{sec:scalability}
 
We now study the complexity of SEM-kMAX, showing that it is linear in the input size. 
Let us focus on a single iteration of the procedure. Each iteration is composed of three different steps: parent set exploration, k-MAX, and data imputation. 

Recall that $n$, $k$ and $d$ represent respectively the number of variables, the bound on the treewidth and the number of data points in the data set. 
As for the first step, the size of the search space of possible parent sets for each variable is ${{n}\choose{k}}$ (without loss of generality, the search can be restricted to parent sets with a size up to the chosen treewidth); for the evaluation of each parent set, we can assume that a scan of the entire dataset is required. This would lead to an overall complexity of $O( nd{{n}\choose{k}})$ for the first step. However, such a complete exploration of the search space of parent sets is usually unfeasible; the algorithm BIC* (\cite{scanagatta2015a}) was designed exactly to overcome this complexity blowup by guiding the exploration to the most promising parent sets in the allowed time ($n$ seconds in our choice). This is equivalent to allowing the exploration of a maximum number of parent sets $p$,  which yields a complexity equal to $ O(ndp)$. Since $p$ is a fixed bound in practice, the actual cost of the first step is $ O(nd)$.
 
 The second step is the execution of k-MAX. This is an iterative algorithm. In every iteration, we first build a core structure of $k$ nodes. Since $k$ is constant by choice, the related computation takes constant time too. The subsequent part of the algorithm adds one node at a time; for each node we explore the pre-computed parent sets, whose maximum number is $p$, which is a constant as we mentioned before. Therefore the overall complexity of one iteration is $O(n)$---note that this part does not depend on data, which have been processed in the first step. The number of iterations is then bounded by the allowed time (in our case $n/10$ seconds) and for this reason it is a constant as well. It follows that the complexity of the second step is dominated by that of the first step.
 
The last step is data imputation, which amounts to performing an exact inference on the bounded-treewidth BN for every record containing missing values. In general the number of queries will then be $O(d)$. The cost of performing each query is $O(n\bar{v}^k)$, where $\bar{v}$ is the maximum number of states for a variable in the network. In our application $k$ is a constant by definition, and $\bar{v}$ can be regarded as a constant too, given that it usually does not exceed few tens in applications. The complexity cost of a query is then $O(n)$, and the final cost of performing data imputation is then $O(nd)$.
 
So far we have considered a single iteration of SEM-kMAX. The number of iterations is usually not very large and in any case can be chosen up to a certain maximum amount. Taking this into account, the overall complexity of SEM-kMAX is $O(nd)$ or, in other words, SEM-kMAX has worst-case complexity linear in the input size (the data). 
 
  \subsection{Parameter tuning}
  \label{subsec:tuning}
 
   \begin{figure}[!ht]
\centering 

\includegraphics{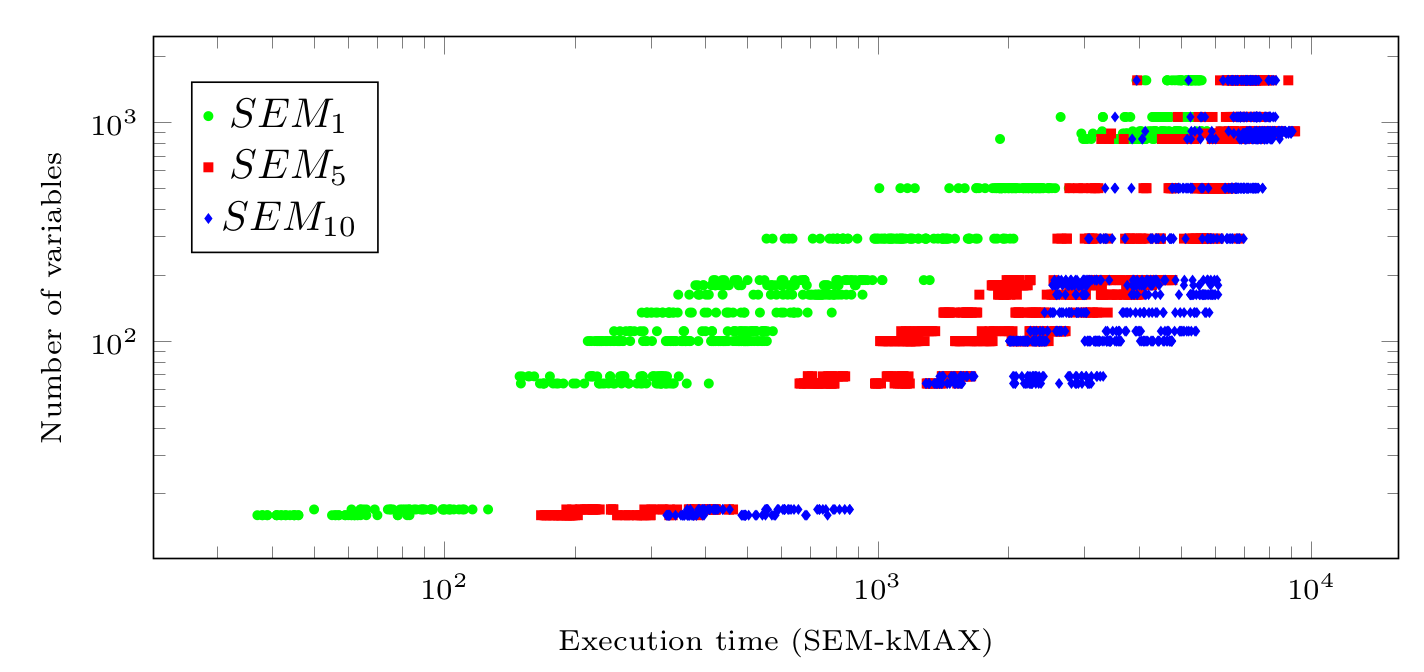}
		
		\caption{Execution time of SEM-kMAX in relation to the number of variables in the dataset. }		
		\label{fig:times-t}
\end{figure}

 As mentioned, one can tune SEM-kMAX by choosing the allowed maximum execution time for its steps. We will now experimentally evaluate different possibilities as for this choice.  
   \begin{figure}[!ht]$SEM_1$
\centering
	
		\begin{subfigure}[b]{0.9\textwidth}
\includegraphics{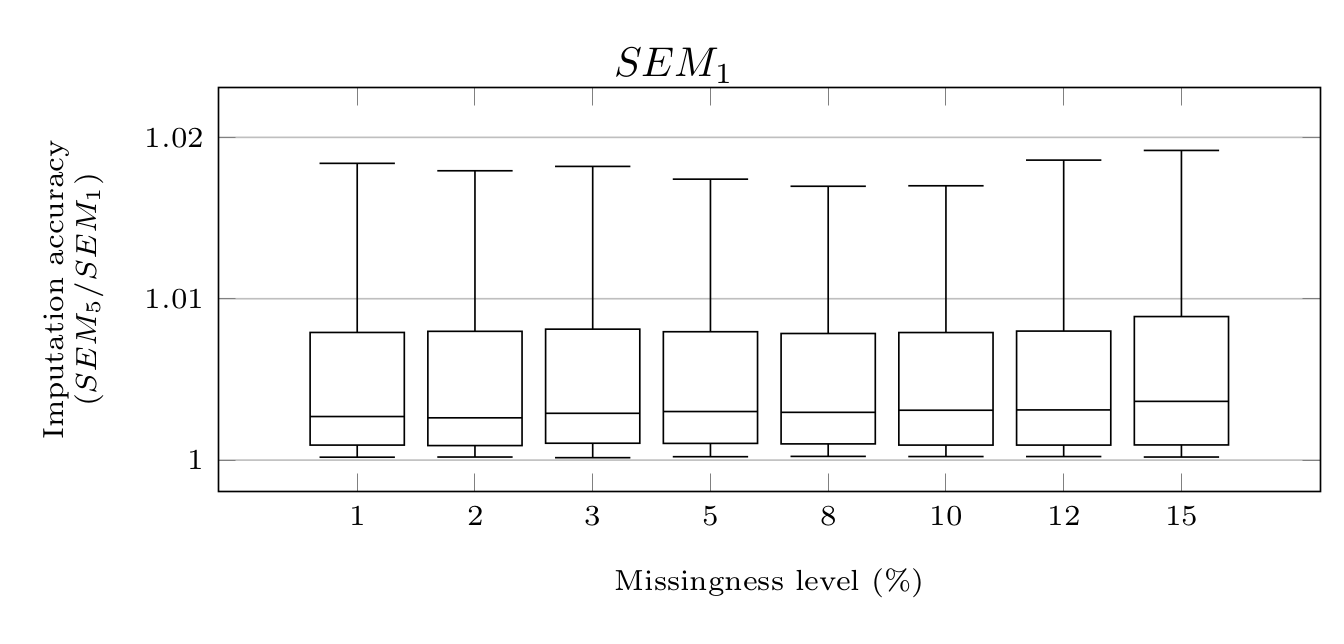}
  	\subcaption{Aggregate comparison of the ratio of imputation accuracy between $SEM_5$ and $SEM_1$.}
  	\label{fig:acc-t-1}
	\end{subfigure}

			\begin{subfigure}[b]{0.9\textwidth}
\includegraphics{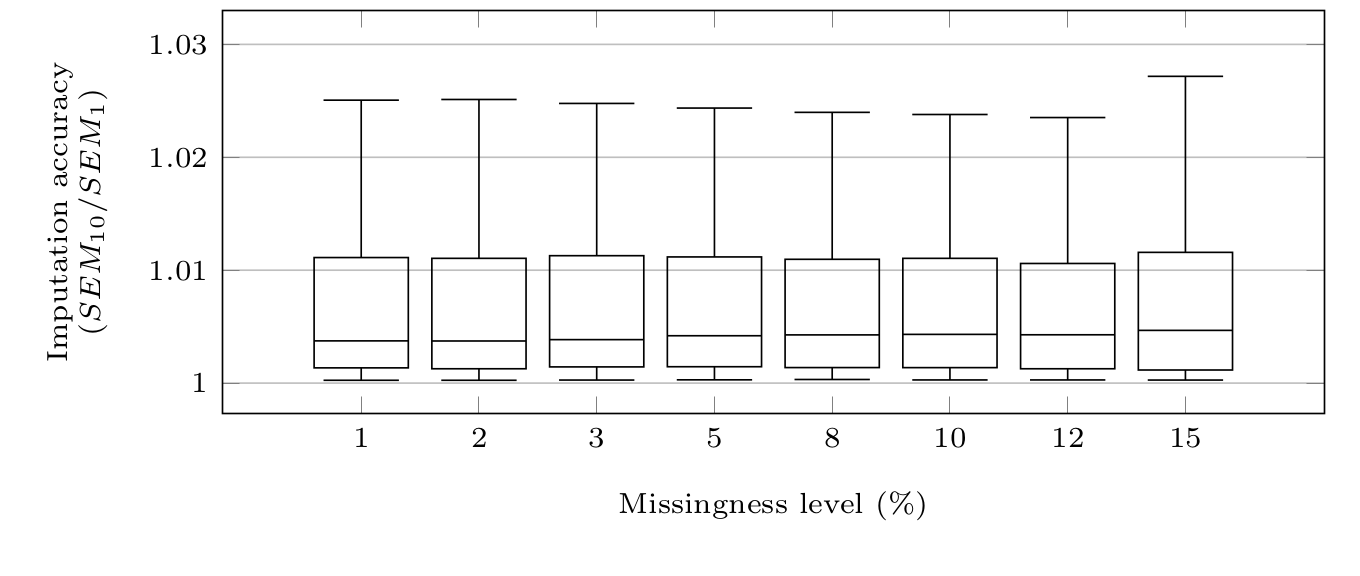}
	\subcaption{Aggregate comparison of the ratio of imputation accuracy between $SEM_{10}$ and $SEM_1$.}  	
	\label{fig:acc-t-2}
	\end{subfigure}
	\caption{Experimental comparison between $SEM_1$, $SEM_5$ snd $SEM_{10}$.}	 	
 \end{figure}
% We can define the effect of this choice as depending on a parameter $t$: 
 We allow for the parent set indentification step $n * t$ seconds, and for the structure optimization step $(n * t) / 10$ seconds, comparing the choices $t = 1$, $t = 5$ and $t = 10$.
 We denote respectively by $SEM_1$,  $SEM_5$ and  $SEM_{10}$  
 the resulting variants of structural EM
 (in the previous experiments we adopted   $SEM_1$,
which is a good compromise between imputation accuracy and required time).
% We denote this configuration as $SEM_1$ (assuming that k-MAX is adopted for structure learning). We additionally executed the experiments on the same data sets with the values of $t = 5$ ($SEM_5$) and $t = 10$ ($SEM_{10}$). 
% Again we considered all the available datasets. We now compare their performances. 

 In Figure~\ref{fig:times-t} we plot the total execution time for all the datasets, for the three configurations. For $SEM_1$ the required time is almost linear in the size of the dataset, showing that learning from incomplete data is feasible even for massive datasets with more than thousands of variables. For $SEM_5$ and $SEM_{10}$ we see that the for the largest dataset the total execution time is closer to  $SEM_1$. The reason for this is the cost for executing the expectation step, which remains almost the same in all the cases. 
 
 From the overall behaviour we can see that the complexity increase in the required execution time is at most linear in the number of variables. 
 In Figure~\ref{fig:acc-t-1} we plot the imputation accuracy of $SEM_5$ against $SEM_1$ and in Figure~\ref{fig:acc-t-2} between $SEM_{10}$ against $SEM_1$. The advantage of $SEM_5$ over $SEM_1$ is clear, while the advantage of $SEM_{10}$ over $SEM_5$ can be called into question. 
 
 By choosing a value for the parameter $t$, the user can tune the trade-off between imputation accuracy and total execution time. We recommend using $SEM_1$ as it strikes an optimal balance  between the two objectives, at least in our experiments. 
% reasonable compromise between required time for the computation and learning accuracy

 \subsection{Parallelization}
 \label{subsec:parallelization}
 
The whole learning procedure can greatly benefit from parallelization. In the parent set identification step, each variable can be considered independently from the others. In the structure optimization k-MAX can be run simultaneously on multiple cores, taking into consideration only the best structures found for each of them. Finally in the expectation phase the queries required for the estimation of the missing values can be executed independently for each data point.

\section{Conclusions}
\label{sec:concl}

We presented a new anytime algorithm (k-MAX) for learning bounded-treewidth Bayesian networks.
Experiments on complete data sets show that k-MAX
finds structures with significantly higher fit to the data than
its competitors, especially on high-dimensional data sets.
Moreover, k-MAX can be plugged within structural EM in order to perform structure learning from incomplete data sets; in this case it allows to efficiently compute the expectation phase thanks to the bounded treewidth.
%We compare the performance of the resulting approach in data imputation.
Structural EM with k-MAX achieves comparable accuracy to state-of-the-art  imputation approaches based on random forest,
while allowing for a speedup of about one order of magnitude.
To the best of our knowledge, our approach is the first implementation of structural EM able to efficiently scale to thousands of variables.

\section{Acknowledgments}
The research in this paper has been partially supported by the Swiss NSF grants n.~IZKSZ2\_162188. 
This work was also supported by the National Research Foundation of Korea(NRF) funded by the Ministry of Science, ICT and Future Planning (NRF-2015K1A3A1A14021055).
We thank Cassio de Polpo Campos for critical discussions on the topics of this paper.

\end{document}